\documentclass[conference,compsoc]{IEEEtran}
\usepackage[english]{babel}
\usepackage{url}  
\usepackage{hyperref}
\ifCLASSOPTIONcompsoc
  \usepackage[nocompress]{cite}
\else
  \usepackage{cite}
\fi

\ifCLASSINFOpdf
   \usepackage[pdftex]{graphicx}
   \graphicspath{{images/}}
\else
  \usepackage[dvips]{graphicx}
  \graphicspath{{images/}}
\fi
\usepackage[cmex10]{amsmath}
\usepackage{listings}
\usepackage{color}
\usepackage{array}

\definecolor{codegreen}{rgb}{0,0.6,0}
\definecolor{codegray}{rgb}{0.5,0.5,0.5}
\definecolor{codepurple}{rgb}{0.58,0,0.82}
\definecolor{backcolour}{rgb}{0.95,0.95,0.92}
 
\lstdefinestyle{mystyle}{
    backgroundcolor=\color{backcolour},   
    commentstyle=\color{codegreen},
    keywordstyle=\color{magenta},
    numberstyle=\tiny\color{codegray},
    stringstyle=\color{codepurple},
    basicstyle=\footnotesize,
    breakatwhitespace=false,         
    breaklines=true,                 
    captionpos=b,                    
    keepspaces=true,                 
    numbers=left,                    
    numbersep=5pt,                  
    showspaces=false,                
    showstringspaces=false,
    showtabs=false,
    tabsize=2
}
 
\usepackage{hyperref}
\hypersetup{
    colorlinks=true,
    linkcolor=blue,
    filecolor=magenta,      
    urlcolor=cyan,
}
 
\begin{document}
%
\title{Benchmarking Traditional Machine Learning and Deep Learning Models for Fault Detection in Power Transformers}

\author{
\IEEEauthorblockN{BHUVAN SARAVANAN\\
\IEEEauthorblockN{107123020@nitt.edu\\National Institute of Technology,\\Tiruchirappalli, TamilNadu, India\\
}
\and
\IEEEauthorblockN{PASANTH KUMAR M D}
\IEEEauthorblockN{107123075@nitt.edu\\National Institute of Technology,\\Tiruchirappalli, TamilNadu, India\\}
}
\and
\IEEEauthorblockN{AARNESH VENGATESON}
\IEEEauthorblockN{107123002@nitt.edu\\National Institute of Technology,\\Tiruchirappalli, TamilNadu, India\\}
}
\maketitle

\footnotetext{|The source code used to implement the proposed model is publicly available at: \url{https://github.com/BhuvanSaravanan/power_transformer_fault_detection}}

\begin{abstract}
Accurate diagnosis of power transformer faults is essential for ensuring the stability and safety of electrical power systems. This research presents a comparative analysis of  conventional machine learning (ML) algorithms and deep learning (DL) algorithms for fault classification of power transformer. Leveraging the condition monitored dataset which span for 10 months, various gas concentration features were normalized and used to train five ML classifiers—Support Vector Machine (SVM), k-Nearest Neighbors (KNN), Random Forest (RF),  XGBoost, and Artificial Neural Network (ANN) —as well as three DL models, including Long Short-Term Memory (LSTM), Gated Recurrent Unit (GRU), One-Dimensional Convolutional Neural Network (1D-CNN), and TabNet. Experimental results demonstrate both conventional machine learning (ML) algorithms and deep learning (DL) algorithms performed at par: the highest accuracy among ML models was achieved by RF at 86.82\%, whereas DL model 1D-CNN emerged very close at 86.30\%. 
\end{abstract}


%
\IEEEpeerreviewmaketitle

\section{Introduction}
Power transformers are indispensable components in electrical power systems, serving as critical nodes for voltage regulation and energy distribution. Their operational integrity directly influences the stability and reliability of the entire power grid. However, transformers are susceptible to various faults due to factors such as insulation degradation, mechanical stresses, thermal overloads, and environmental conditions. These faults can lead to catastrophic failures, resulting in substantial economic losses and power outages\cite{cao2024advancement}.

Traditional diagnostic techniques, including Dissolved Gas Analysis (DGA), Partial Discharge (PD) monitoring, and Infrared Thermography (IRT), have been employed to detect and analyze transformer faults. While these methods have proven effective in identifying certain fault types, they often fall short in providing comprehensive, real-time diagnostics. For instance, DGA requires periodic oil sampling and laboratory analysis, which may delay fault detection \cite{chakravorti2013recent}. PD monitoring is sensitive to external noise and may not accurately localize fault sources. IRT primarily detects surface temperatures and may miss internal anomalies.

The limitations of conventional diagnostic methods have prompted the exploration of advanced techniques that leverage artificial intelligence (AI) and machine learning (ML) algorithms. These intelligent approaches can process vast amounts of data, identify complex patterns, and provide predictive insights into transformer health. For example, AI-based models have been developed to enhance the accuracy of fault classification by analyzing multiple parameters simultaneously, such as gas concentrations, temperature profiles, and vibration signals\cite{madan1997applications}.

Moreover, the integration of multi-source data has emerged as a promising direction in transformer fault diagnosis. By combining electrical and non-electrical parameters, such as oil and gas characteristics, vibration signals, and thermal images, multi-source data fusion techniques can offer a holistic view of transformer conditions. This comprehensive analysis enables more accurate fault detection and prognosis, facilitating proactive maintenance strategies and reducing the risk of unexpected failures\cite{zhang2021adaptive}.

In summary, the evolution of transformer fault diagnosis is moving towards intelligent, data-driven methodologies that overcome the shortcomings of traditional techniques. By harnessing the capabilities of AI and multi-source data integration, these advanced diagnostic systems aim to enhance the reliability, efficiency, and safety of power transformers within the electrical grid.

\section{Literature Survey}

Power transformers are critical components in electrical power systems, and their reliable operation is paramount. Traditional diagnostic techniques have been employed for decades to monitor and assess the health of these transformers. Among these, Dissolved Gas Analysis (DGA), Partial Discharge (PD) monitoring, and Infrared Thermography (IRT) are the most prevalent.

\textbf{Dissolved Gas Analysis (DGA):} DGA is a widely used method for detecting incipient faults in oil-immersed transformers. It involves analyzing the types and concentrations of gases dissolved in transformer oil, which are byproducts of thermal and electrical faults. Various interpretation methods, such as the Key Gas Method, Duval Triangle, and Rogers Ratio, have been developed to diagnose specific fault types based on gas concentrations\cite{sun2012review}. The Duval Triangle, introduced in the 1970s, utilizes the relative percentages of methane (CH\textsubscript{4}), ethylene (C\textsubscript{2}H\textsubscript{4}), and acetylene (C\textsubscript{2}H\textsubscript{2}) to classify faults into categories like partial discharges, low-energy discharges, and high-energy discharges\cite{ali2023conventional}. Despite its effectiveness, DGA has limitations, including the need for periodic oil sampling and potential delays in fault detection.

\textbf{Partial Discharge (PD) Monitoring:} PD monitoring detects localized dielectric breakdowns within the transformer insulation system, which can be precursors to major failures. Techniques such as ultra-high frequency (UHF) sensors and acoustic emission detectors are employed to capture PD signals. However, PD detection is sensitive to external noise, and accurate localization of PD sources can be challenging. Moreover, the installation of PD monitoring equipment can be complex and costly, limiting its widespread adoption\cite{ali2023conventional}.

\textbf{Infrared Thermography (IRT):} IRT is a non-contact diagnostic technique that captures thermal images of transformer surfaces to identify hotspots indicative of abnormal heating. It is particularly useful for detecting issues like loose connections, overloading, and cooling system failures. Recent advancements have integrated IRT with machine learning algorithms to enhance fault classification accuracy . Nevertheless, IRT primarily detects surface temperatures and may not reveal internal faults. Additionally, environmental factors such as ambient temperature and emissivity can affect measurement accuracy\cite{mahami2022automated}.

While these traditional diagnostic methods have been instrumental in transformer maintenance, they each have inherent limitations. DGA requires manual sampling and may not provide real-time data. PD monitoring systems can be expensive and susceptible to interference. IRT offers surface-level insights but may miss internal anomalies. These constraints highlight the need for more advanced, integrated diagnostic approaches that provide comprehensive, real-time monitoring to ensure transformer reliability and longevity.

\section{Methodology  }
\subsection{Data Description}
Power transformers, integral to the stability and efficiency of modern power grids, require continuous monitoring to avoid catastrophic failures. Despite their generally robust construction, transformers are susceptible to both \textbf{mechanical} and \textbf{dielectric failures}, which may result from physical stress, insulation degradation, or electrical overloads. These failure mechanisms often manifest subtly, making early detection a critical element of predictive maintenance systems.

In recent years, the integration of \textbf{Internet of Things (IoT)} devices has enabled the collection of fine-grained operational data for real-time transformer health assessment. This study uses a data set originally curated for fault detection research on power transformers \cite{putchala2022transformer}. The dataset spans from \textbf{June 25, 2019, to April 14, 2020}, with time-stamped readings captured at \textbf{15-minute intervals}. The data comprises a wide range of sensor inputs capturing vital transformer conditions such as \textbf{voltage, current, oil temperature, winding temperature}, and ambient environmental metrics.

 The dataset includes sensor readings from multiple critical points of the transformer infrastructure. These sensors monitor both electrical and thermal properties that are essential for identifying fault signatures. Table 1 summarizes the key sensors and their abbreviations, adapted from the original documentation in \cite{putchala2022transformer}

\begin{table}[htbp]
\centering
\caption{Abbreviations and Full Sensor Name}
\label{tab:sensor_abbreviations}
\begin{tabular}{|c|l|} \hline 

\textbf{Abbreviations} & \textbf{              Sensor Name}\\ \hline 

VL1  & Phase Line 1 \\ \hline  
VL2  & Phase Line 2 \\ \hline  
VL3  & Phase Line 3 \\ \hline  
IL1  & Current Line 1 \\ \hline  
IL2  & Current Line 2 \\ \hline  
IL3  & Current Line 3 \\ \hline  
VL12 & Voltage Line 1 2 \\ \hline  
VL23 & Voltage Line 2 3 \\ \hline  
VL31 & Voltage Line 3 1 \\ \hline  
INUT & Neutral Current \\ \hline  
OTI  & Temperature Indicator \\ \hline  
WTI  & Winding Temperature Indicator \\ \hline  
ATI  & Ambient Temperature Indicator \\ \hline  
OLI  & Oil Level Indicator \\ \hline  
OTI\textsubscript{A}  & Oil Temperature Indicator Alarm \\ \hline  
OTI\textsubscript{Tr} & Oil Temperature Indicator Trip \\ \hline 
MOG\textsubscript{A}  & Magnetic Oil Gauge Indicator \\ \hline

\end{tabular}
\end{table}

These sensors were selected due to their direct influence on identifying potential transformer stress or degradation. For example, \textbf{OTI and WTI} help detect thermal anomalies that may arise from insulation failure or excessive electrical loading. Similarly, \textbf{OLI and MOG\textsubscript{A}} provide early warnings related to oil loss or leakage, which can compromise transformer insulation and cooling.

\subsection{Fault Annotation and Label Construction}

In order to enable the application of machine learning models, especially those suited for binary classification tasks, a fault labeling mechanism was introduced. Specifically, readings from the following binary fault indicators were used to define the target class:

\begin{itemize}
    \item \textbf{WTI} (Winding Temperature Indicator)
    \item \textbf{OTI\textsubscript{A}} (Oil Temperature Alarm)
    \item \textbf{OTI\textsubscript{T}} (Oil Temperature Trip)
    \item \textbf{MOG\textsubscript{A}} (Magnetic Oil Gauge Alarm)
\end{itemize}
If any of the above sensors signaled a fault (value of 1), the transformer instance was labeled as \textbf{"Faulty"}. Conversely, if all fault-related sensors reported normal readings (value of 0), the instance was marked as \textbf{"Healthy."} This labeling approach enables downstream binary classification modeling, transforming the dataset into a supervised learning framework.

To reduce redundancy, the original four binary fault indicator columns were dropped after the “Faulty Transformer” label was generated. This ensures that the input features do not leak target information during training, thereby preserving model integrity.

 \section{Evaluation Strategy}

All models were evaluated on a balanced dataset using an 80/20 train-test split. Standard classification metrics were used to assess model performance:

\begin{itemize}
    \item \textbf{Accuracy}: Overall proportion of correct predictions
    \item \textbf{Precision}: Correct positive predictions as a percentage of all positive predictions
    \item \textbf{Recall}: Proportion of actual faults correctly identified
    \item \textbf{F1-Score}: Harmonic mean of precision and recall
\end{itemize}

This metric suite ensures fair evaluation, particularly for imbalanced datasets where models may be biased toward the majority class.
\\
\textbf{All deep learning models were trained using the Adam optimizer with a learning rate of 0.001 and binary cross-entropy loss. The models were evaluated on the test set using the same metrics as the conventional models. }

\section{Conventional Machine Learning Models }

The dataset used in this study was preprocessed to ensure high-quality input for the models. Non-numeric columns were removed, and the remaining features were scaled using the \verb|StandardScaler| to normalize the data. To address class imbalance, the Synthetic Minority Oversampling Technique (SMOTE) was applied, generating a balanced dataset for training and testing 

\subsection{Random Forest} 

Random Forest is an ensemble of decision trees that operates by training multiple trees and aggregating their outputs through majority voting. It is particularly robust to overfitting and performs well with structured, tabular data such as transformer telemetry. In the baseline study, Random Forest demonstrated the highest overall classification performance.
\textbf{A robust ensemble method with multiple decision trees, optimized for the number of estimators, maximum depth, and other parameters. }

\subsection{Support Vector Machine (SVM)}

SVM constructs hyperplanes in a high-dimensional space to separate the binary classes with a maximum margin. Kernel tricks allow it to handle non-linearly separable data. SVM is known for its generalization ability, particularly on small to medium datasets. \textbf{A radial basis function (RBF) kernel was used, and hyperparameter were optimized using grid search. }

\subsection{K-Nearest Neighbors (KNN)}

KNN is a non-parametric model that classifies instances based on the majority label among the \textit{k} closest samples in the feature space. It is simple and interpretable but sensitive to noisy data and high dimensionality. \textbf{A distance-based classifier with hyperparameter tuning for the number of neighbors and weighting schemes. }

\subsection{XGBoost}

XGBoost (Extreme Gradient Boosting) is a scalable, tree-based ensemble algorithm that uses gradient boosting to iteratively improve model accuracy. It handles missing values internally and has built-in regularization to prevent overfitting. \textbf{A gradient-boosting framework optimized for learning rate, maximum depth, and the number of estimators}. 

\subsection{Artificial Neural Network (ANN)}

A feedforward artificial neural network with dense layers is used to model complex non-linear relationships in the input data. Despite requiring more training time and tuning, ANNs are capable of capturing intricate patterns that simpler models may overlook.  \textbf{multi-layer perceptron with a single hidden layer of 100 neurons, trained for 300 iterations. }

\begin{table}[!htbp]
\centering
\caption{Performance of Conventional Machine Learning Models}
\begin{tabular}{lcccc}
\hline
\textbf{Model} & \textbf{Accuracy} & \textbf{Precision} & \textbf{Recall} & \textbf{F1 Score} \\
\hline
Random Forest & 0.8682 & 0.8042 & 0.9780 & 0.8826 \\
SVM           & 0.8604 & 0.7933 & 0.9800 & 0.8768 \\
KNN           & 0.8624 & 0.8129 & 0.9463 & 0.8745 \\
XGBoost       & 0.8680 & 0.8045 & 0.9769 & 0.8824 \\
ANN           & 0.8632 & 0.8030 & 0.9674 & 0.8776 \\
\hline
\end{tabular}
\end{table}

\section{Deep Learning Models}

\subsection{Long Short-Term Memory (LSTM)}
Long Short-Term Memory (LSTM) networks are a type of recurrent neural network (RNN) designed to effectively capture long-range temporal dependencies in sequential data. In the context of power transformer fault diagnosis, LSTM networks are particularly suited to model time-series data such as voltage, current, or thermal sensor readings. Their internal memory gates (input, forget, and output gates) enable them to retain relevant historical information while discarding irrelevant data, allowing for robust modeling of fault patterns that evolve over time. LSTM has demonstrated strong performance in a variety of industrial time-series classification tasks, making it a logical choice for this application.\textbf{ A recurrent neural network (RNN) variant designed to handle sequential data. The model consisted of 64 LSTM units followed by a dense layer with a sigmoid activation function. }

\subsection{Gated Recurrent Unit (GRU) }
Gated Recurrent Units (GRUs) are a simplified variant of LSTM networks that also model sequential data but with fewer parameters and faster training times. GRUs utilize update and reset gates to control the flow of information, enabling them to capture temporal dependencies while avoiding issues such as vanishing gradients. For transformer fault diagnosis, GRUs offer a computationally efficient alternative to LSTMs, especially when working with long sequences or limited computational resources, without significantly compromising accuracy. \textbf{Another RNN variant, similar to LSTM but with fewer parameters, making it computationally efficient}. 

\subsection{One-Dimensional Convolutional Neural Network (1D-CNN) }
One-Dimensional Convolutional Neural Networks (1D-CNNs) are particularly effective for extracting local temporal patterns from time-series data. By applying convolutional filters across input sequences, 1D-CNNs automatically learn discriminative features related to sudden changes or periodic fluctuations in operational parameters that may indicate a fault. Their relatively simple architecture and low training cost make them suitable for rapid deployment in industrial fault detection systems. Moreover, 1D-CNNs can be stacked or combined with recurrent layers to enhance their temporal modeling capacity. \textbf{A convolutional model with a single convolutional layer, followed by a dense layer. This model was designed to extract spatial features from the input data}. 

\subsection{TabNet }
TabNet is a deep learning architecture specifically designed for tabular data, which often characterizes power system monitoring datasets. It utilizes a sequential attention mechanism to select relevant features at each decision step, enabling both interpretability and high predictive performance. Unlike traditional dense neural networks, TabNet can learn feature interactions dynamically and focuses on sparsity, leading to efficient learning and better generalization. In transformer fault diagnosis, TabNet’s ability to highlight influential features can aid both in fault classification and in understanding underlying failure mechanisms.\textbf{ A deep learning model specifically designed for tabular data, leveraging attention mechanisms to focus on important features.} 

\begin{table}[!htbp]
\centering
\caption{Performance of Deep Learning Models}
\begin{tabular}{lcccc}
\hline
\textbf{Model} & \textbf{Accuracy} & \textbf{Precision} & \textbf{Recall} & \textbf{F1 Score} \\
\hline
LSTM    & 0.8491 & 0.7964 & 0.9435 & 0.8637 \\
GRU     & 0.8560 & 0.7920 & 0.9710 & 0.8724 \\
1D-CNN  & 0.8630 & 0.7966 & 0.9800 & 0.8788 \\
TabNet  & 0.8393 & 0.7606 & 0.9969 & 0.8628 \\
\hline
\end{tabular}
\end{table}

\section{Results and Discussion }

The performance of both conventional machine learning models and deep learning models was evaluated on the task of predicting transformer faults. The results were assessed using standard metrics, including accuracy, precision, recall, and F1-score. This section discusses the findings and provides insights into the performance of the models.

\subsection{Results Of Conventional Machine Learning Models}

The conventional machine learning models included Support Vector Machine (SVM), Random Forest, K-Nearest Neighbors (KNN), XGBoost, and Artificial Neural Network (ANN). These models were trained on a balanced dataset created using SMOTE to address class imbalance. The features were scaled using \verb|StandardScaler| to ensure uniformity across all input dimensions.

Among the conventional models:

\begin{itemize}
    \item \textbf{Random Forest} achieved the highest performance across all metrics, with an accuracy of 86.82\%, precision of 80.42\%, recall of 97.80\%, and an F1-score of 88.26\%. Its ability to handle complex feature interactions and its robustness to overfitting contributed to its superior performance.
    \item \textbf{XGBoost} also performed well, with an accuracy of 86.80\% and an F1-score of 88.24\%. The ensemble nature of Random Forest allowed it to generalize effectively, though it slightly lagged behind XGBoost in precision and recall.
    \item \textbf{ANN} demonstrated competitive performance with an accuracy of 86.32\%, but its recall was lower compared to ensemble methods, indicating potential challenges in identifying minority class instances.
    \item \textbf{KNN} showed moderate performance, with an accuracy of 86.24\%. Its reliance on distance metrics made it sensitive to feature scaling and the choice of hyper-parameters.
    \item \textbf{SVM}, despite being a shallow neural network, achieved an accuracy of 86.04\%, showcasing its ability to model non-linear relationships in the data.
\end{itemize}

\subsection{Results Of Deep Learning Models}

The deep learning models included Long Short Term Memory (LSTM), Gated Recurrent Unit (GRU), 1D Convolutional Neural Network (1D-CNN), and TabNet. \textbf{These models were trained for 20 epochs using the Adam optimizer with a learning rate of 0.001. The binary cross-entropy loss function was used to optimize the models, and the data was reshaped into a 3D format for sequential models.}

Among the deep learning models:

\begin{itemize}
    \item \textbf{1D-CNN} emerged as the best-performing model, achieving an \textbf{accuracy of 86.30\%}, \textbf{precision of 79.66\%}, \textbf{recall of 98\%}, and an \textbf{F1-score of 87.88\%}. Its convolutional layers effectively captured spatial patterns in the data.
    \item \textbf{GRU} achieved an accuracy of 85.60\% and an F1-score of 79.20\%.Its reduced parameter count made it computationally efficient while maintaining competitive performance.
    \item \textbf{LSTM} performed similarly to LSTM, with an accuracy of 84.91\% and an F1-score of 86.37\%. Its ability to capture temporal dependencies in the data contributed to its strong performance, though it required more computational resources compared to other models.
    \item \textbf{TabNet} achieved an accuracy of 83.93\% and an F1-score of 86.28\%. and recall of 99.69\%.Its attention mechanism allowed it to focus on the most relevant features, making it particularly effective for tabular data.
\end{itemize}

\subsection{Analysis of ROC Curves and AUC for Model Performance}

The Receiver Operating Characteristic (ROC) curves presented in the figure provide a comprehensive evaluation of the classification performance of multiple machine learning and deep learning models applied to the dataset. The ROC curve plots the True Positive Rate (TPR) against the False Positive Rate (FPR) at various threshold settings, offering a visual representation of the trade-off between sensitivity and specificity for each model. The Area Under the Curve (AUC) is a scalar value summarizing the model's ability to distinguish between the positive and negative classes, with higher values indicating better performance.

\begin{figure}[h]
    \centering
    \includegraphics[width=1\linewidth]{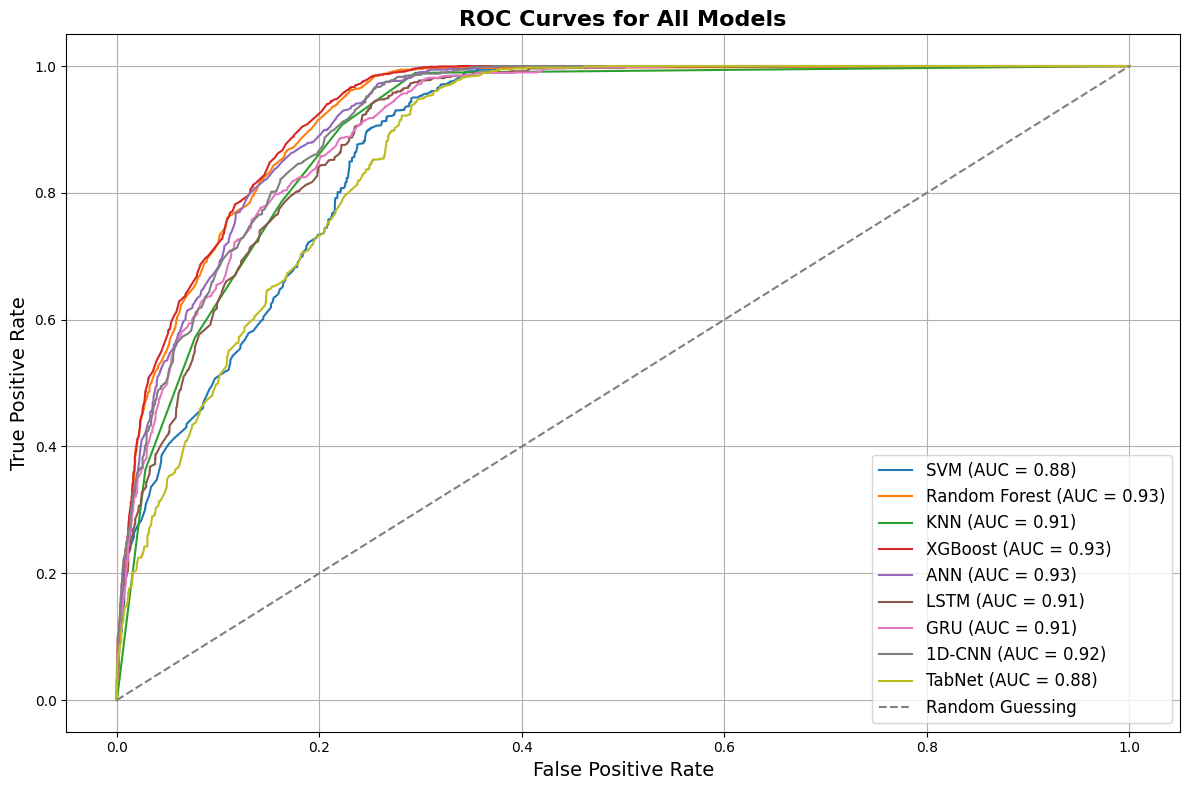} 
    \caption{Receiver Operating Characteristic (ROC) curves  }
    \label{fig:example}
\end{figure}

\subsection{Key Observations:}

 \begin{itemize}

\item \textbf{Random Forest, XGBoost, and ANN}:
These models exhibit the highest AUC values (approximately 0.93), indicating superior classification performance. Their ROC curves are closer to the top-left corner of the plot, suggesting a strong ability to maximize TPR while minimizing FPR. This performance can be attributed to their inherent ability to capture complex patterns in the data through ensemble learning (Random Forest and XGBoost) and multi-layered neural networks (ANN).

\textbf{\item Deep Learning Models (LSTM, GRU, and 1D-CNN)}:
The LSTM, GRU, and 1D-CNN models achieve AUC values around 0.91–0.92, demonstrating competitive performance. These models are particularly effective in handling sequential and structured data, which may explain their strong results. The slight variation in AUC among these models could be due to differences in their architectures and the way they process temporal dependencies.

\textbf{\item KNN and SVM}:
The K-Nearest Neighbors (KNN) and Support Vector Machine (SVM) models achieve AUC values of approximately 0.88–0.91. While these models perform well, their slightly lower AUC values compared to ensemble and deep learning models suggest that they may not capture the underlying data complexity as effectively. SVM's performance is likely enhanced by the use of the radial basis function (RBF) kernel, which handles non-linear decision boundaries.

\textbf{\item TabNet}:
The TabNet model achieves an AUC of 0.88, which, while lower than the top-performing models, still reflects a reasonable classification capability. TabNet's performance may be influenced by its reliance on attention mechanisms and its ability to handle tabular data efficiently.

  \end{itemize}

\section{Limitations and Future Work }

\subsection{Limitations}

\textbf{Feature Engineering}:
The study relied on automated feature scaling and balancing techniques (e.g., SMOTE) but did not incorporate domain-specific knowledge into feature engineering. Incorporating expert knowledge about transformer operations and fault mechanisms could improve model interpretability and performance.
   
\textbf{Imbalanced Dataset Challenges}:
Although SMOTE was used to address class imbalance, synthetic oversampling may introduce noise into the dataset. Alternative techniques, such as adaptive synthetic sampling (ADASYN) or cost-sensitive learning, could be explored to mitigate this issue

 \subsection{Future Works}

\textbf{Temporal and Spatial Analysis}:
The study primarily focused on tabular data. Future work could incorporate temporal and spatial data, such as time-series sensor readings or geographic transformer locations, to improve fault prediction accuracy. Techniques like spatiotemporal modeling or graph neural networks (GNNs) could be explored.

\textbf{Economic and Environmental Impact}:
Future studies could evaluate the economic and environmental impact of deploying these models. For instance, quantifying the cost savings from early fault detection or the reduction in carbon emissions due to optimized transformer maintenance schedules.

\newpage


\bibliographystyle{IEEEtran}
\bibliography{reference}
%



\end{document}